\documentclass[sigconf]{acmart}
\usepackage{booktabs} 
\usepackage[ruled]{algorithm2e} 
\usepackage{balance} 
\begin{document}
\title{Multimodal Continuous Turn-Taking Prediction Using Multiscale RNNs}

\author{Matthew Roddy}
\affiliation{%
  \institution{ADAPT Centre, Trinity College}
  \city{Dublin}
  \country{Ireland}
}
\email{roddym@tcd.ie}

\author{Gabriel Skantze}
\affiliation{%
  \institution{Speech Music and Hearing, KTH}
  \city{Stockholm}
  \country{Sweden}
}

\author{Naomi Harte}
\affiliation{%
  \institution{ADAPT Centre, Trinity College}
  \city{Dublin}
  \country{Ireland}
}

\copyrightyear{2018} 
\acmYear{2018} 
\setcopyright{acmcopyright}
\acmConference[ICMI '18]{2018 International Conference on Multimodal Interaction}{October 16--20, 2018}{Boulder, CO, USA}
\acmBooktitle{2018 International Conference on Multimodal Interaction (ICMI '18), October 16--20, 2018, Boulder, CO, USA}
\acmPrice{15.00}
\acmDOI{10.1145/3242969.3242997}
\acmISBN{978-1-4503-5692-3/18/10}

\begin{abstract}
In human conversational interactions, turn-taking exchanges can be coordinated using cues from multiple modalities. To design spoken dialog systems that can conduct fluid interactions it is desirable to incorporate cues from separate modalities into turn-taking models. We propose that there is an appropriate temporal granularity at which modalities should be modeled. We design a multiscale RNN architecture to model modalities at separate timescales in a continuous manner. Our results show that modeling linguistic and acoustic features at separate temporal rates can be beneficial for turn-taking modeling. We also show that our approach can be used to incorporate gaze features into turn-taking models. 
\end{abstract}



\maketitle
\section{Introduction}

The design of naturalistic spoken dialog systems (SDSs) that can interact with users in a human-like manner, requires models that control when it is an appropriate time for the system to speak. These turn-taking models (e.g. \cite{raux_finite-state_2009,ferrer_is_2002,maier_towards_2017}) are used to make decisions as to whether the current speaker will continue to hold the floor (HOLD) or not (SHIFT). A naive approach to modeling these decisions is to identify turn exchange points using silence thresholding.  This approach is limited in its naturalness since the chosen threshold can potentially be too short (leading to interruptions) or too long (leading to unnaturally long pauses). These traditional threshold-based approaches also cannot model rapid turn-switches, which can potentially involve a brief period of overlap during the turn exchange \cite{heldner_pauses_2010}.

It has been proposed in the projection theory of Sacks \cite{sacks_simplest_1974} that, rather than reacting to silence, humans often form predictions as to when their conversational partner will finish their turn. Using these predictions we are able to anticipate utterance endpoints, and start our turns accordingly. LSTM-based turn-taking models that operate in a similar predictive fashion were proposed by Skantze in \cite{skantze_towards_2017}. In these models LSTMs are used to make continuous predictions of a person's speech activity at each individual time step of 50ms. The networks are trained to predict a vector of probability scores for speech activity in each individual frame within a set future window. Rather than designing classifiers to make specific decisions, these continuous models are able to capture general information about turn-taking in the data that they are trained on. They can therefore be applied to a wide variety of turn-taking prediction tasks and have been shown to outperform traditional classifiers when applied to HOLD/SHIFT predictions.

A downside to the approach in \cite{skantze_towards_2017} is that, since a single LSTM is being used, all input features must be processed at the same rate. When considering linguistic features, the the relevant information for turn-taking prediction happens at a coarse temporal granularity in comparison with acoustic features, where much of the relevant information occurs at the sub-word prosodic level. When using a single LSTM, this requires either averaging the acoustic features to represent them at a word-level temporal resolution, or upsampling the linguistic features to represent them at the acoustic temporal resolution. Both of these options have their drawbacks. In the case of averaging the acoustic features, we lose the finer-grained prosodic inflections that are important to forming turn-taking predictions. In the case of upsampling the linguistic features, the model has a harder time with the longer term dependencies that exist between linguistic features. Because the linguistic features are sampled at a higher rate, the model is more susceptible to the vanishing gradient problem. We propose a way to address this problem by using a multiscale RNN architecture, in which modalities are modeled in separate sub-network LSTMs that are allowed to operate at their own independent timescales. A master LSTM is used fuse the modalities and form predictions at a regular rate by taking as input a concatenation of the current hidden states of the sub-networks. This allows the hidden states of the sub-networks to be updated at an appropriate temporal rate. 

In this paper we present significant extensions to the original work of Skantze in \cite{skantze_towards_2017}. We investigate the performance of our proposed multiscale architecture on two datasets that contain two different combinations of modalities. We look at the influence of modeling modalities in separate sub-networks and using separate timescales. We find that there are significant performance benefits to modeling linguistic features at a slower temporal rate, and in a separate sub-network from acoustic features. We also find that our approach can be used to incorporate gaze features into turn-taking models, a task that has been previously found to be difficult  \cite{de_kok_multimodal_2009}.

\section{Continuous Turn-taking Prediction}

The main objective of continuous turn-taking prediction as proposed in \cite{skantze_towards_2017} is to predict the future speech activity annotations of one of the speakers in a dyadic conversation using input speech features from both speakers ($x_t$). At each timestep $t$ of frame-size 50ms, speech features are extracted and input into an LSTM that is used to predict the future speech activity ($y_t$) of one of the speakers. The future speech activity is a 3 second window comprising of 60 frames of the binary annotations for frames $t+1$ to $t+60$. The output layer of the network uses an element-wise sigmoid activation function to predict a probability score for the target speaker's speech activity at each future frame. To represent linguistic features in this model the word-rate features are upsampled to the 50ms acoustic feature rate. They use one-hot encodings, where the feature is "switched on" for a single frame, 100ms after the word is finished (to simulate real-world ASR conditions). To model more fine-grained prosodic inflections the 50ms acoustic feature rate could be increased. However, doing this leads to a sparser linguistic feature representation since all features input into an LSTM must be processed at the same rate. This makes it more difficult for the network to model longer-term dependencies that exist in the linguistic modality.

\begin{figure*}[ht]
\begin{center}
  \makebox[\textwidth]{\includegraphics[width=0.92\textwidth]{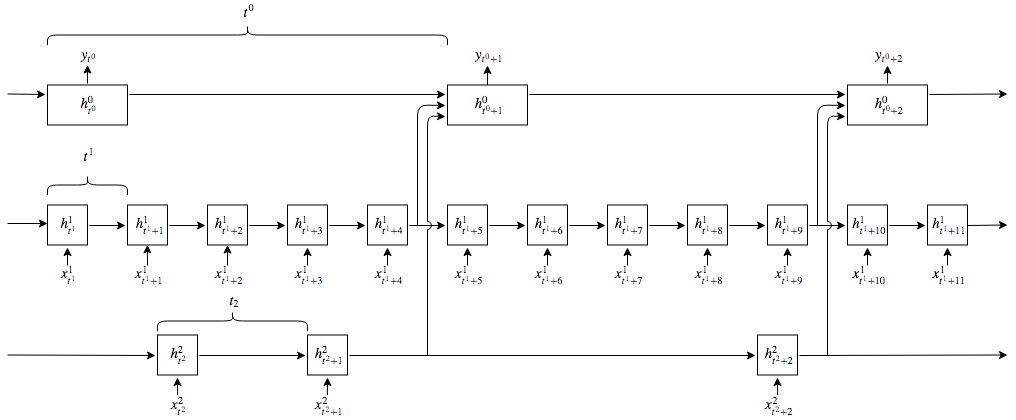}}
  \caption{Multiscale RNN architecture }
  \label{fig:multiscale}
\end{center}
\end{figure*}
\section{Multiscale Continuous Turn-taking Prediction}

To address this problem we modify the original network architecture to include a variable number of sub-network LSTM cells that process features from separate modalities independently. The sub-networks are allowed to process the input features from the separate modalities at different timescales. An example network configuration that uses a master LSTM ($h^0$) and two sub-network LSTMs ($h^1$,$h^2$, each assigned a modality) is shown in Fig. \ref{fig:multiscale}. We use superscripts to denote the index of modalities $m\in M$, and subscripts to index timesteps (represented using the notation $t^m$). At each timestep of the master LSTM ($t^0$), the current states of the sub-network LSTMs are concatenated and fed into the master LSTM. The hidden state update process for the network is shown in Algorithm \ref{multiscale_algo}. 
By feeding the current states of the sub-networks into the master LSTM, we are effectively performing a sampling operation, represented in the algorithm by the step $h_{t^0+1}^m  \longleftarrow h_{t^m+1}^m$. The sampling operation can either increase or decrease the temporal resolution of the individual modalities, depending on the timescales used. For example, in Fig. \ref{fig:multiscale}, the temporal resolution of the first sub-network ($h^1$) is decreased since we sample it at a regular rate every five $t^1$ timesteps. The temporal resolution of the second modality could either be increased or decreased by the sampling process since the features have irregular timesteps. The processing of features at a slower update rate potentially allows the network to better retain information. The model was implemented using the PyTorch framework and our code is available online\footnote{www.github.com/mattroddy/lstm\_turn\_taking\_prediction}.

\begin{algorithm}[ht]
\SetAlgoLined
 \KwIn{ $h_{t},$ $x_{t+1}$  } 
 \KwOut{ $y_{t^0+1}$ }
 \For{$m \in M $}{
    \For{$t^m: {t^0}\leq t^m\leq t^0+1$ }{
        $h_{t^m+1}^m \longleftarrow LSTM(x_{t^m+1}^m,h_{t^m}^m;\Theta^m)$
      }
    $h_{t^0+1}^m  \longleftarrow h_{t^m+1}^m$
 }
 $h_{t^0+1}^0 \longleftarrow LSTM([ h_{t^0+1}^m,..., {h_{t^0+1}^M}]^T, h_{t^0}^0;\Theta^0)$ \;
 $y_{t^0+1} \longleftarrow \sigma(h_{t^0+1}^0;\Theta^\sigma)$ 
 \caption{Multiscale continuous turn-taking prediction}
 \label{multiscale_algo}
\end{algorithm}

\section{Experimental Design}
To assess the performance of our multiscale approach, we test it on two different datasets. In each dataset, features from two separate modalities are investigated by training models using a variety of different network configurations. The HCRC map-task corpus (MTC) \cite{anderson_hcrc_1991} is used to examine linguistic and acoustic modalities while the Mahnob Mimicry Database (MMD) \cite{bilakhia_mahnob_2015} is used to examine visual and acoustic modalities. In this section we discuss the details of the datasets and how features were extracted. We then discuss the evaluation metrics used to assess network performance. We then outline our experimental procedure.
\subsection{Map-Task corpus}
The MTC is a corpus of 128 dyadic task-based dialogues totaling 18 hours in length. We used 96 conversations as training and 32 conversations for testing. Additionally, we used 32 conversations of the training set as a held-out test set during hyperparameter searches.  The speech transcriptions supplied with the corpus were used as the ground-truth for the speech-activity predictions.
\paragraph{Acoustic Features}\label{acous_feats}
As acoustic features, we use the low-level descriptors from the eGeMAPs \cite{eyben_geneva_2016} feature set extracted with the OpenSmile toolkit \cite{eyben_opensmile_2010}. The features were all normalized using z-scores on a per-file basis. We extract the features at two different temporal resolutions: 10ms and 50ms. We use these two different temporal resolutions to investigate which one is more useful for our turn-taking models. In our results tables and discussion we refer to these as "Acous 10ms" and "Acous 50ms".  
\paragraph{Linguistic Features}\label{ling_section}
For linguistic features we use the word annotations supplied with the corpus. 
The words were represented as an enumerated vocabulary where the raw word features were transformed into a linear embedding of size 64 that is jointly trained with the rest of the network. 
In an effort to simulate the conditions of a real-time system, the linguistic features were not provided to the system until 100ms after the end of the word. Three different temporal rates for the processing of linguistic features are tested in our experiments. In our discussion and results below, "Ling 50ms" refers to using word features that have been sampled at regular 50ms intervals, as was proposed in \cite{skantze_towards_2017}. "Ling 10ms" refers to using word features that are sampled at a faster rate of 10ms. "Ling Asynch" refers to using an irregular update rate, where the LSTM only processes the linguistic features when a new word is available.

\subsection{Mahnob Mimicry Database}
The MMD is an audio-visual corpus of 54 dyadic conversations totaling 11 hours in length. The participants are either assigned discussion topics or roles to play in a loosely defined role-playing scenario. When splitting the data into training and test sets we balanced the number of role-playing and discussion conditions in the training and test set. We used 39 conversations for training and 15 for testing. Since there are no speech transcriptions available for the dataset, we manually labeled the dataset for speech activity. The procedure we used for extracting acoustic for MMD was the same as that followed in for the MTC in section \ref{acous_feats}. 
\paragraph{Visual Features}
We automatically extract visual features using the OpenFace toolkit \cite{baltrusaitis_openface_2016}. During informal exploratory experiments we found that the automatically extracted gaze features performed better than other features extracted with the toolkit (e.g. facial action units, pose). We therefore used the gaze features (a six dimensional vector of eye gaze directions) along with a confidence score as our visual input feature. The video in the MMD uses a high frame rate of 58Hz. We perform a comparison of using features at this high frame rate and using features that are averaged over 50ms frame windows. In the results tables and the discussion below we refer to the high frame rate video and the averaged video features as "Visual 58Hz" and "Visual 50ms" respectively.

\subsection{Evaluation Metrics}

To evaluate the performance of the different network configurations we use two kinds of evaluation metrics that were proposed in \cite{skantze_towards_2017}. These two kinds of metrics represent the prediction performance on two common types of turn-taking decisions that are pertinent to SDSs.  

\paragraph*{Prediction at Pauses}

The prediction at pauses metric represents the standard turn-taking decision made at brief pauses in the interaction to predict whether the person holding the floor will continue speaking (HOLD) or the interlocutor will take a turn (SHIFT). To make this decision, we find all points where there is a pause of a minimum set length. We then select all of these instances where only one person continued within a one second window directly after the pause. We average the predicted output probabilities within the window for each of the speakers at the frame directly after the pause. The speaker with the higher score is selected as the predicted next speaker, giving us a binary HOLD/SHIFT classification for which we report F-scores. We test predictions at pauses of both 500ms (PAUSE 500) and 50ms (PAUSE 50). The majority vote F-score for the MTC (always HOLD) is 0.5052 and 0.4608 for PAUSE 50 and PAUSE 500 respectively. For MMD the corresponding values are 0.7298 and 0.8185.

\paragraph*{Prediction at Onsets}

Prediction at onsets (ONSET) represents a prediction of the length of an utterance after an initial period of speech. It represents a useful decision for estimating how long the upcoming utterance will be. It categorizes onset predictions into SHORT and LONG, where SHORT utterances can be considered similar to backchannels. For an utterance to be classified as short, 1.5 seconds of silence by a participant has to be followed by a maximum of 1 second of speech, after which the speaker must be silent for at least 5 seconds. For the utterance to be classified as long, 1.5 seconds of silence must be followed by at least 2.5 seconds of speech. The point at which the predictions are made is 500ms from the start of the utterance. The prediction is made by taking the mean of the 60 output nodes from the sigmoid layer and comparing them to a threshold value. The majority vote F-score for the MTC (always SHORT) is 0.3346. For the MMD the corresponding value is 0.5346. 
\begin{table}[t]
\centering
\caption{Map-task corpus experimental results}
\label{tab:tab_maptask}
\resizebox{\columnwidth}{!}{

\begin{tabular}{lrrrr}
\toprule 
 & BCE loss  & f1\ 50ms  & f1\ 500ms  & f1\ onset \tabularnewline
\midrule
No Subnets (Early Fusion) &  &  &  & \tabularnewline
\midrule
(1) Acous 50ms  & 0.5456  & 0.7907  & 0.8165  & 0.7926 \tabularnewline
(2) Acous 10ms  & 0.5351  & 0.8154  & 0.8428  & 0.8126 \tabularnewline
(3) Ling 50ms  & 0.5779  & 0.7234  & 0.7547  & 0.7249 \tabularnewline
(4) Ling Asynch  & 0.5839  & 0.7101  & 0.7341  & 0.7174 \tabularnewline
(5) Ling 10ms  & 0.5823  & 0.7072  & 0.7391  & 0.7111 \tabularnewline
(6) Acous 50ms Ling 50ms  & 0.5411  & 0.7957  & 0.8354  & 0.8101 \tabularnewline
(7) Acous 10ms Ling 10ms  & \textbf{0.5321}  & \textbf{0.8194}  & \textbf{0.8465}  & \textbf{0.8141} \tabularnewline
\midrule 
One Subnet &  &  &  & \tabularnewline
\midrule
(8) Acous 50ms Ling 50ms  & 0.5414  & 0.7922  & 0.8366  & 0.8020 \tabularnewline
(9) Acous 10ms Ling 10ms  & \textbf{0.5317 } & \textbf{0.8237 } & \textbf{0.8480 } & \textbf{0.8128 }\tabularnewline
\midrule 
Two Subnets (Multiscale) &  &  &  & \tabularnewline
\midrule
(10) Acous 50ms Ling 50ms  & 0.5420  & 0.7916  & 0.8303  & 0.8019 \tabularnewline
(11) Acous 10ms Ling 50ms  & \textbf{\emph{0.5291}}  & \textbf{\emph{0.8323}}  & 0.8526  & \textbf{\emph{0.8236}} \tabularnewline
(12) Acous 50ms Ling Asynch  & 0.5416  & 0.7949  & 0.8385  & 0.7993 \tabularnewline
(13) Acous 10ms Ling Asynch  & 0.5296  & 0.8307  & \textbf{\emph{0.8553}}  & 0.8232 \tabularnewline
(14) Acous 10ms Ling 10ms  & 0.5310  & 0.8285  & 0.8470  & 0.8189 \tabularnewline
\bottomrule
\end{tabular}

}

\end{table}

\begin{table}[t]
\centering
\caption{Mahnob corpus experimental results}
\label{tab:tab_mahnob}
\resizebox{\columnwidth}{!}{

\begin{tabular}{lrrrr}
\toprule 
{}  & BCE loss  & f1\ 50ms  & f1\ 500ms  & f1\ onset \tabularnewline
\midrule 
No Subnets (Early Fusion) &  &  &  & \tabularnewline
\midrule 
(1) Acous 50ms  & 0.4433  & 0.8665  & 0.9230  & 0.8668 \tabularnewline
(2) Acous 10ms  & \textbf{0.4348}  & \textbf{\emph{0.8851}}  & \textbf{0.9343}  & \textbf{0.8685} \tabularnewline
(3) Visual 50ms  & 0.5840  & 0.7858  & 0.8154  & 0.6445 \tabularnewline
(4) Visual 58Hz  & 0.5941  & 0.7726  & 0.8031  & 0.6560 \tabularnewline
(5) Acous 50ms Visual 50ms  & 0.4497  & 0.8651  & 0.9159  & 0.8526 \tabularnewline
\midrule
Two Subnets (Multiscale) &  &  &  & \tabularnewline
\midrule
(6) Acous 50ms Visual 50ms  & 0.4443  & 0.8637  & 0.9198  & 0.8711 \tabularnewline
(7) Acous 10ms Visual 50ms  & 0.4337  & \textbf{0.8840 } & \textbf{\emph{0.9347}}  & \textbf{\emph{0.8784}} \tabularnewline
(8) Acous 50ms Visual 58Hz  & 0.4437  & 0.8634  & 0.9216  & 0.8721 \tabularnewline
(9) Acous 10ms Visual 58Hz  & \textbf{\emph{0.4332}}  & 0.8831 & 0.9343  & 0.8762 \tabularnewline
\bottomrule
\end{tabular}

}
\end{table}

\subsection{Experimental Procedure}
In our experiments, the networks were trained to minimize binary cross entropy (BCE) loss which was shown to produce good results in \cite{roddy_investigating_2018}. We test the impact of using three different network configurations with multiple combinations of modalities at different temporal resolutions. The three network configurations are: "no subnets", which corresponds to an early fusion approach in which the modalities are fed directly into a single LSTM; "one subnet", which corresponds to the use of only one sub-network LSTM; and "two subnets", which corresponds to the use of separate LSTM sub-networks for the individual modalities. We note that combinations such as "Ling 50ms" with "Acous 10ms" are not possible when using the "no subnets" and "one subnet" configurations since the features are being input into the same LSTM and cannot operate at different temporal resolutions. Grid searches for three hyperparameters (hidden node size, dropout, and L2 regularization) were performed for each network configuration. In order to limit the influence of parameter count changes between the different network configurations, the hidden node count in a given network was limited to a sum of 150. Once the hyperparameters for a network are chosen, we train the network five times and report the mean values of the different evaluation metrics in Tables \ref{tab:tab_maptask} and \ref{tab:tab_mahnob}. The best performing modality combination for a given network configuration is shown in bold and the best overall performance is shown in italics. In our discussion below we use two-tailed t-tests to report on the difference between the means of metrics.  

\section{Discussion}

Looking at the results from the fusion of linguistic and acoustic modalities shown in Table \ref{tab:tab_maptask}, it is clear that there are significant benefits in modeling acoustic and linguistic modalities separately using different timescales. Our best performance on all evaluation metrics is achieved using our multiscale approach where features from the two modalities are modeled at separate rates. Comparing the BCE loss of the best performing early-fusion result (7) with the best multiscale result (11) gives a statistically significant improvement ($P<.001$). Comparing the performance of the acoustic feature timescales, we observe that the faster rate of 10ms consistently performs better than the slower 50ms rate. Looking at the performance of the three different linguistic timescales in (3,4,5), we see that processing linguistic features at the slower regular rate of 50ms achieves the best performance. Comparing the BCE loss of (3) and (5) suggests that sampling linguistic features at a fast temporal rate makes it difficult for the network to model longer term dependencies ($P=.004$). The effect of processing modalities on their own in separate sub-networks without the added gain of using separate timescales is inconclusive when we examine (6) and (10). Using a single subnet as an added layer also does not yield significant differences to the early fusion approach. We conclude that the main advantage in using our multiscale approach on a combination of acoustic and linguistic modalities is its ability to fuse the two modalities when the linguistic features are operating at a slow 50ms timescale and the acoustic features are operating at a fast 10ms timescale. Comparing our results with previously published baselines reported on the same dataset by Skantze in \cite{skantze_towards_2017}, our best result on the PAUSE 500 task of 0.8553 is a large improvement over his reported score of 0.762. 

Looking at the results from the fusion of visual and acoustic modalities shown in in Table \ref{tab:tab_mahnob}, we were able to achieve our best BCE loss using our multiscale approach to fuse acoustic features at a 10ms timescale and visual features at a 58Hz timescale. Comparing this result (9) with our best "no subnets" result (2) gives a statistically significant improvement ($P=0.035$). We note that using early fusion with gaze features (5) does not add any value when compared to acoustic features on their own (1). The results also indicate that the faster 58Hz gaze features perform better than the averaged 50ms visual features when used in conjunction with the acoustic features. This suggests that we loose relevant information by averaging the gaze features within a timestep.

\section{Conclusion}
In conclusion, we have shown that there are considerable benefits in using multiscale architectures for continuous turn-taking prediction. When fusing linguistic and acoustic modalities, the architecture allows acoustic features to be modeled at a fast temporal rate (which is better-suited to capturing prosodic inflections) while modeling linguistic features at a slower rate (which is better-suited to capturing long-term linguistic dependencies). When fusing visual and acoustic modalities, our multiscale approach allowed the use of high frame-rate visual features without resorting to averaging. 

\section{Acknowledgements}
\scriptsize{ The ADAPT Centre for Digital Content Technology is
funded under the SFI Research Centres Programme (Grant
13/RC/2106) and is co-funded under the European Regional
Development Fund.}

\bibliographystyle{ACM-Reference-Format}
\balance{}
\bibliography{My_Library.bib}

\end{document}